\documentclass{article}

\usepackage[final]{corl_2021} % Uncomment for the camera-ready ``final'' version.
\usepackage[compact]{titlesec}
\usepackage{times}
\usepackage{epsfig}
\usepackage{graphicx}
\usepackage{amsmath}
\usepackage{amssymb}
\usepackage{wrapfig}

\usepackage{booktabs}       % professional-quality tables
\usepackage{array}
\usepackage{multirow}
\usepackage{color}
\usepackage[normalem]{ulem}
\usepackage{mathtools}
\usepackage{resizegather}

\usepackage{caption}
\usepackage{url}
\usepackage{epsfig}
\usepackage{multirow}

\usepackage{xcolor}
\usepackage{comment}
\usepackage{bm}

\DeclareMathOperator*{\argmin}{argmin}
\DeclareMathOperator*{\expect}{\mathbb{E}}

\newcommand{\norm}[1]{\left\lVert#1\right\rVert}
\newcommand\figref{Figure~\ref}
\newcommand\tabref{Table~\ref}

\definecolor{forestgreen}{rgb}{0.133, 0.545, 0.133}
\definecolor{brown}{rgb}{0.65, 0.16, 0.16}

\newcommand{\ourdataset}{\textsc{Xenith}}

\makeatletter
\def\blfootnote{\gdef\@thefnmark{}\@footnotetext}
\makeatother

\title{Exploring Adversarial Robustness of Multi-sensor Perception Systems in Self Driving}

\author{
    $^{1,2}$~James Tu%\thanks{Denotes the correspondence author (james.tu@mail.utoronto.ca).}
    \quad
    $^{3}$~Huichen Li%\thanks{Work done while an intern at Uber ATG.}
    \quad
    Xinchen Yan
    \quad
    $^{1,2}$~Mengye Ren
    \quad
    $^{1,2}$~Yun Chen
    \and
    $^{1}$~\textbf{Ming Liang
    \quad
    $^{4}$~Eilyan Bitar %\thanks{The contribution to this publication was carried out as consultant to Uber ATG.}
    \quad
    Ersin Yumer
    \quad
    $^{1,2}$~Raquel Urtasun}
    \vspace{1mm}
    \and
    Waabi$^1$, University of Toronto$^2$, UIUC$^3$, Cornell University$^4$ \and
\texttt{\footnotesize \{jtu, mren, yun, urtasun\}@cs.toronto.edu}
\quad
\and
\texttt{\footnotesize huichen3@illinois.edu}
\quad
\texttt{\footnotesize eyb5@cornell.edu}
\and
\texttt{\footnotesize \{skywalkeryxc,liangming.elgoog,meyumer\}@gmail.com}
\quad
}

\titlespacing{\section}{3pt}{*0}{*0}
\begin{document}
\maketitle
\blfootnote{Work done while all authors were at UberATG.} 

%===============================================================================
\begin{abstract}
Modern self-driving perception systems have been shown to improve upon processing complementary inputs such as LiDAR with images. In isolation, 2D images have been found to be extremely vulnerable to adversarial attacks.
Yet, there are limited studies on the adversarial robustness of multi-modal models that fuse LiDAR and image features. 
Furthermore, existing works do not consider physically realizable perturbations that are consistent across the input modalities.
In this paper, we showcase practical susceptibilities of multi-sensor detection by inserting an adversarial object on a host vehicle. We focus on physically realizable and input-agnostic attacks that are feasible to execute in practice, 
and show that a single universal adversary can hide different host vehicles from state-of-the-art multi-modal detectors.
Our experiments demonstrate that successful attacks are primarily caused by easily corrupted image features. 
Furthermore, in modern sensor fusion methods which project image features into 3D, adversarial attacks can  exploit  the projection process to generate false positives in distant regions in 3D.
Towards more robust multi-modal perception systems, we show that adversarial training with feature denoising can boost robustness to such attacks significantly. 
%However, we find that standard adversarial defenses still struggle to prevent false positives which are also caused by inaccurate associations between 3D LiDAR points and 2D pixels.
\end{abstract}

% Two or three meaningful keywords should be added here
\keywords{Adversarial, Self-Driving, Perception, Multimodal} 

%===============================================================================

%!TEX root = main.tex

\section{Introduction}\label{sec:intro}
Recent advances in self-driving perception have shown that fusing information from  multiple sensors  (e.g., camera,  LiDAR, radar)~\cite{gupta2014learning,song2016deep,wang2018depth,qi2018frustum,liang2019multi,radarnet} leads to superior performance when compared to relying on single sensory inputs.
Such performance gains are primarily due to the complementary information contained in the measurements provided by the different types of sensors.
For example, LiDAR sensors provide accurate 3D geometry while cameras capture rich appearance information. 

%\begin{figure}[t]
%\centering
%\includegraphics[width=0.7\linewidth]{figs/teaser}
%\cutcaptionup
%\vspace{-6pt}
%\caption{
%In this work, we generate a universal adversarial mesh to hide various host vehicles from state-of-the-art multi-sensor object detectors.
%
%Our attack produces consistent perturbations across image and LiDAR modalities.
%}
%\vspace{-3pt}
%\cutcaptiondown
%\label{fig:method}
%\end{figure}

Meanwhile, modern perception models which rely on deep neural networks (DNNs) have been found to be extremely vulnerable to adversarial attacks when processing images in isolation~\cite{eykholt2018robust,xie2017adversarialdetect,lu2017adversarial,phystexture,ranjan2019attacking}. 
Adversarial attacks can be thought of as perturbations to the sensory inputs which do not alter the semantic meaning of the scene, but drastically change a DNN's output and resulting in incorrect predictions. 
Such vulnerabilities can lead to catastrophic consequences in safety-critical applications.
In the context of self-driving, most efforts have investigated attacks against single-sensor inputs, such as image-only attacks~\cite{eykholt2018robust,ranjan2019attacking} and LiDAR-only attacks~\cite{tu2020physically}. Towards multi-modal robustness, \cite{wang2020towards} considers perturbations of LiDAR and image inputs independently, resulting in perturbations that are inconsistent across modalities and therefore may not be physically realizable and hence not threatening in practice.
On the other hand, some proposed physically realizable approaches \cite{cao2020msf} only search over shape but ignore texture which is crucial for corrupting image inputs. 
Furthermore, these prior works do not generate universal perturbations which are perhaps the most threatening in practice. 
Such perturbations are input agnostic and can 
%successfully 
attack any input in the training distribution with high probability, meaning they can be executed without prior knowledge of the scene and are able to consistently disrupt models that process sensory information across time.

\begin{figure*}[t]
\centering
\includegraphics[width=1.0\linewidth]{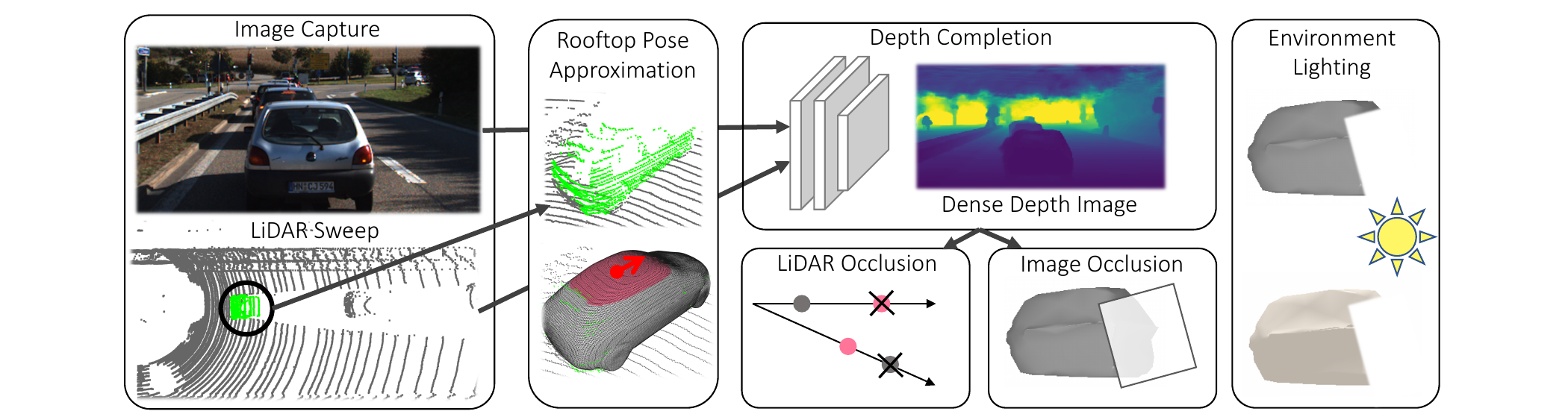}
\caption{Simulating the addition of a mesh onto a vehicle rooftop in a realistic manner. First roof approximation is done to determine placement location and heading. Then LiDAR points and pixels are rendered with directional lighting to approximate sunlight. Finally, a dense depth image is generated with depth completion and used to handle occlusion.}
\vspace{-8mm}
\label{fig:sim}
\end{figure*}

This paper demonstrates the susceptibility of multi-sensor detection models to physically realizable and input-agnostic adversarial perturbations. 
To create a physically realizable attack which is also feasible to execute, we focus on object insertion attacks~\cite{eykholt2018robust,athalye2018synthesizing,xiao2019meshadv,xiang2019generating,tu2020physically}, as they can be carried out via the deployment of physical objects in the real world.
Following \cite{tu2020physically}, we insert the adversarial object by placing it on the rooftop of a host vehicle.
We render the   adversary into LiDAR and image inputs to ensure perturbations are consistent across modalities and that our attack is physically realizable.
Furthermore, we consider occlusion and environmental lighting in the rendering process as shown in \figref{fig:sim} to enhance the realism of simulation. 
To enable end-to-end learning of geometry and texture, we render the pixels and LiDAR points in a differentiable manner.
During training, our adversary is optimized with respect to all vehicles in the training distribution to create a universal attack which can be applied to any vehicle in any scene. 

We conduct an empirical evaluation of our proposed attack on the KITTI~\cite{Geiger2012CVPR} 
self-driving dataset and a novel large-scale self-driving dataset \ourdataset{} using the multi-sensor detector MMF~\cite{liang2019multi}. 
We generate input-agnostic adversarial examples that successfully hide host vehicles from state-of-the-art detectors in both datasets. 
More importantly, we find that incorporating image inputs makes models more vulnerable when compared to using LiDAR alone, as successful attacks are primarily caused by the brittle image features.
Moreover, the projection of image features into 3D allows the adversary to generate false detections in distant regions.
Nonetheless, we show that false negative failures can be circumvented by applying feature denoising and adversarial training. 
However, we observe that distant false positives are much harder to correct with adversarial defense, as they are also caused by inaccurate mappings between 2D pixels and 3D LiDAR points during fusion. 
%

%!TEX root = main.tex
\section{Related work}\label{sec:related}
Adversarial attacks were first discovered in the 2D image domain, where small perturbations on the
pixels were shown to generate drastically different prediction results on object
classification~\cite{szegedy2014intriguing,goodfellow2015explaining}. Networks trained on object
detection and semantic segmentation have also been shown to exhibit such
vulnerability~\cite{xie2017adversarialdetect,lu2017adversarial,chen2018shapeshifter,liu2018dpatch,li2018robust,wei2019transferable}. 
Early
methods~\cite{szegedy2014intriguing,goodfellow2015explaining,deepfool,universal} assume the
perfect knowledge of the gradients of the victim model, referred to as \textit{whitebox} attacks.  Later it was found that a \textit{blackbox} attack can achieve similar success as
well~\cite{practicalblackbox,decisionattack,qeba}.
Defense and robustness evaluation procedures have also been explored for adversarial attacks ~\cite{athalye2018obfuscated,papernot2016limitations,carlini2017towards,ilyas2019adversarial,carlini2019evaluating,tsipras2018robustness,xie2019feature,engstrom2019exploring}.

Aside from changing the pixel values by a small amount, various other ways to ``perturb'' an image
were also proposed. Object insertion attacks are realistic attacks that insert an object
to change the network output while not introducing changes in
semantics~\cite{advpatch,athalye2018synthesizing,eykholt2018robust,yang2020patchattack, hamdi2020sada}.
These attacks were originally designed to be stickers that can be attached to a target object, and has since also been applied to the lens of a camera~\cite{camsticker}. 
%Semantic adversarial
%attacks~\cite{semanticadv,semanticadviccv,bhattad2019unrestricted}, on the other hand, aim to use a
%generative model to edit the semantic properties of an image, while maintaining the original identity. 
Image rendering is another popular technique for non-pixel based attacks, which can
also be made differentiable~\cite{diffrender}, by using which \cite{beyondimage} showed that
adversarial attacks can be made through changing lighting.  Various other object
insertion attacks designed camouflage textures that can be wrapped around the target
object~\cite{camou,phystexture,advcamou,universalcamou,advlogo,wu2020making}.

%The safety and robustness of self-driving cars against adversarial attacks has also been widely studied.
%
Aside from the typical image-based attacks introduced above, adversarial attacks against point clouds have also been studied.
%since LiDAR sensors are usually equipped by self-driving vehicles, a body of work focused on point clouds as the input modality. 
%
\cite{xiang2019generating,advpc,pointsetsadv,hamdi2020advpc} tried to directly perturb the location and cardinality of point clouds. 
However, such attacks may not be physically realizable, as arbitrary perturbations won't be achievable by a LiDAR sensor with fixed angular frequency.
Towards more realistic attacks, \cite{sensorattack,robustlidar} developed spoofing attacks that add malicious LiDAR points, while other approaches~\cite{xiao2019meshadv,cao2019adversarial,tu2020physically} instead optimize 3D mesh surfaces and use differentiable ray-casting to generate LiDAR point clouds.
%Compared to previous works on LiDAR attacks, we
%are more focused on analyzing vulnerabilities of image and
%LiDAR and how adding image inputs affects robustness.
%The inclusion of the image modality is an important novel addition to previous attacks, as multi-sensor models are becoming increasingly popular but their adversarial robustness has not yet been explored.

Despite the fact that multi-modal sensor configurations are widely seen on self-driving
vehicles~\cite{contfuse,liang2019multi,mvxnet,fadadu2020multi,bijelic2020seeing,chadwick2019distant}, research on multi-modal sensor attacks is
still very limited. Several preliminary works show the possibility of attacking multi-sensor fusion
networks~\cite{wang2020towards,cao2020msf,yu2020multimodal}. However, \cite{wang2020towards}
did not consider consistency across data modalities when perturbing the image input, whereas~\cite{cao2020msf} did not
consider image texture, resulting in a lack of attack expressivity, and~\cite{yu2020multimodal} did
neither. 
%Although experiments show that attacking multiple modalities together can contribute to a
%stronger attacks, recent research also shows a seemingly different conclusion that multi-task
%learning can generate more robust networks~\cite{mao2020multitask}. 
%
We believe that it would be an
interesting question to ask, whether multi-sensor fusion can be made more robust when the attacks
are both input-agnostic and physically realizable.

\begin{figure*}[t]
\centering
\includegraphics[width=.82\linewidth]{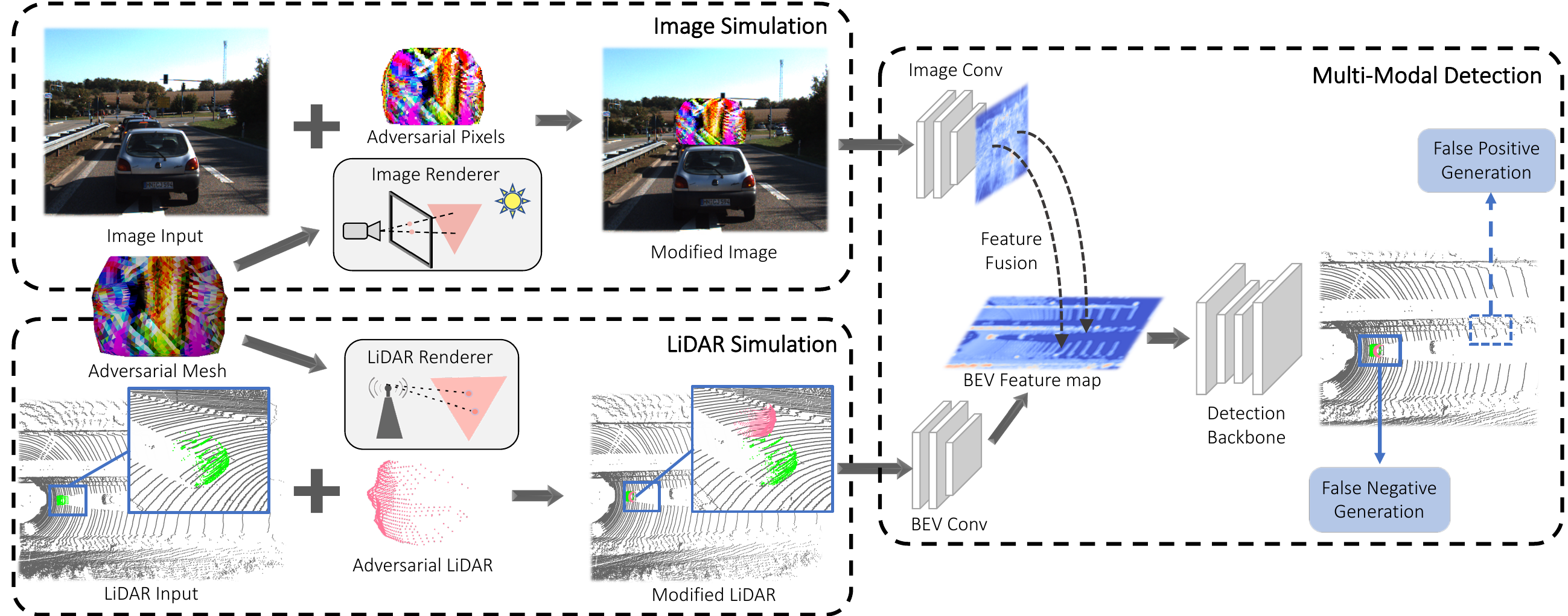}
\caption{Overview of the attack pipeline. The adversarial mesh is rendered into both LiDAR and image inputs in a differentiable manner. 
The inputs are then processed by a multi-sensor detection model which outputs bounding  box proposals. 
An adversarial loss is then applied to generate false negatives by suppressing correct proposals and false positives by encouraging false proposals. 
Since the entire pipeline is differentiable, gradient can flow from the adversarial loss to mesh parameters.}
\vspace{-5mm}
\label{fig:method}
\end{figure*}

\section{Multi-sensor Adversarial Learning}\label{sec:method}
In this section, we present a general method for learning an adversarial textured mesh to attack any multi-sensor object detector that is differentiable end-to-end.
%
%In the following, we focus on the LiDAR-image object detector MMF~\cite{liang2019multi}, a state-of-the-art multi-sensor network used in modern self-driving systems.
%
%The multi-sensor fusion module used in MMF is a common building block for other related works~\cite{fadadu2020multi,bijelic2020seeing,chadwick2019distant,radarnet} and our focus on the MMF model does not affect the generality of our results.
%making our study general to other multi-sensor perception models.
%
Specifically, we require the adversarial attack to be (1) input-agnostic for different environments, (2) geometrically-consistent across image and LiDAR input modalities, and (3) fully-automatic for implementation at large-scale.
Our attacks are focused on vehicles as they are the most common object of interest on the road.
%
%To that end, we design our framework using physically-realizable LiDAR attacks~\cite{tu2020physically,cao2019adversarial}, differentiable rendering techniques~\cite{liu2019soft}, and multi-modal sensor fusion networks~\cite{liang2019multi}.
%

\noindent
\textbf{Preliminaries:}
We consider a bird's eye view (BEV) object detection model $F$  that takes the front camera image $x_\text{I} \in {[0, 1]}^{H \times W \times 3}$ and LiDAR point clouds $x_\text{L} \in \mathbb{R}^{P \times 3}$ as input $x = (x_\text{I}, x_\text{L})$.
Here, the dimensions $H$ and $W$ represent the image height and width respectively.
%
%The dimension 
$P$ is
%represents 
the number of LiDAR points which could vary in each frame.
The object detector is trained on bird's eye view (BEV) bounding box annotations $\mathcal{Y}$, with each bounding box instance $b \in \mathcal{Y}$ parameterized by $b = (b_x, b_y, b_h, b_w, b_\alpha)$.
Subsequently, $b_x$ and $b_y$ are coordinates of the bounding box center, $b_h$ and $b_w$ indicate the width and height, respectively, and $b_\alpha$ represents the orientation.
In order to process both image and LiDAR data modalities, the object detector uses two separate branches to extract features from each modality (see Fig~\ref{fig:method}).
Then, the 2D image features are projected into 3D space to be fused with the LiDAR features.

\subsection{Multi-sensor Simulation for Object Insertion}\label{sec:sensor-sim}
In this work, we design a framework to insert a textured mesh into the scene so that both appearance and shape can be perturbed to attack multi-sensor perception systems.
We attach a triangle mesh $\mathcal{M} = (\mathcal{V}, \mathcal{F}, \mathcal{T})$ onto the roof of a host vehicle, as such placement is physically realizable in the real world.
The mesh is parameterized by vertex coordinates $\mathcal{V} \in \mathbb{R}^{N \times 3}$, vertex indices of faces $\mathcal{F} \in \mathbb{N} ^ {M \times 3}$, and per-face vertex textures $\mathcal{T} \in \mathbb{R} ^ {M \times C \times C \times 3}$.
The dimensions $N$, $M$ and $C$ represent the number of vertices, the number of triangle faces, and the per-face texture resolution, respectively.
For scalability reasons, we do not consider transparency, reflective materials, or shadowing, as handling each case would require sophisticated physics-based rendering.
Instead, we approximate the sensor simulation using LiDAR ray-tracing  and a light-weight differentiable image renderer.
Both image and LiDAR rendering pipelines are differentiable, allowing gradients from LiDAR points and image pixels to flow into the mesh parameters during optimization.
The overall pipeline of %multi-sensor simulation 
for object insertion is illustrated in \figref{fig:sim}.

\noindent
\textbf{Rooftop Approximation:}
First, we estimate the center of the vehicle's rooftop to determine the 3D location for placing the adversary. 
Following ~\cite{engelmann2017samp,tu2020physically,najibi2020dops}, we represent our vehicle objects using signed distance functions (SDFs) and further project them onto a low-dimensional shape manifold using PCA.
For each vehicle, we then optimize the low-dimension latent code that minimizes the fitting error between the vehicle point clouds and the shape manifold.
Then, a fitted SDF is converted to a watertight vehicle mesh with Marching Cubes~\cite{lorensen1987marching}. 
We select the top 20cm of the vehicle as the approximate rooftop and use the rooftop center and vehicle heading to determine the exact pose for object insertion.

\noindent
\textbf{LiDAR Simulation:}
To simulate insertion in the LiDAR sweep, we sample rays according to LiDAR sensor specifications used to collect the original sweep, such as the number of beams and the horizontal rotation rate. 
We then compute the intersection of these rays and mesh faces using the Moller-Trumbore algorithm~\cite{moller1997fast,tu2020physically} to obtain a simulated point cloud of the adversarial mesh. 
These simulated points are then added to the original LiDAR sweep.

\noindent
\textbf{Image Simulation:} 
To render the adversary into the image, we extract the intrinsics and extrinsics from the camera sensor that captured the original image. 
We then use a light-weight differentiable renderer SoftRas~\cite{liu2019soft} to simulate the mesh pixels. 
Using a soft rasterizer during optimization allows gradient flow from occluded and far-range vertices to enable better 3D reasoning from pixel gradients.
To enhance the fidelity of rendered images, we model the sun light with a directional lighting model as a light at infinite distance.

\noindent
\textbf{Occlusion Reasoning:} 
As we insert a new object into the scene, the rendering process must also consider occlusion for both the original and newly rendered points and pixels.
To handle LiDAR occlusions, we compare the depth of existing points in the LiDAR sweep and newly rendered points, discarding the point that is farther away. 
%
%For each ray which generates a point in the original LiDAR sweep and on the adversarial mesh, we take the closer point and discard the other point which has been occluded.
%Image occlusion can be handled in a similar manner
Unlike LiDAR where the depth of each point is known, raw images do not contain depth information.
To obtain depth estimates, first the LiDAR points are projected onto the image to generate a sparse depth image since images have higher resolution than LiDAR.
We then use a depth completion model~\cite{chen2019learning}, which uses the sparse depths and RGB image to compute dense per-pixel depth estimates.
%
%Using the dense depth map, we discard rendered pixels which have greater depth than the corresponding pixel in the original image.
%
With depth estimates of each pixel, we overlay rendered pixels onto the original image and discard occluded pixels.
Note that we do not attack the depth completion model as it is a preprocessing step used to increase the fidelity of simulation.

\subsection{Universal Adversarial Example Generation}
\label{sec:multi-adv-attack}
%With differentiable rendering methods that simulate consistent perturbations across image and LiDAR input modalities,
%we now describe the process to generate multi-sensor adversarial examples to fool the object detection model $F$, parametrized by $\theta$.
%\xcyan{, parametrized by $\theta$.}
%\xcyan{Reflecting R1 comment ``$\theta$ of Eq (8) should be included in Eq (2)'.'}
%
%We denote the detection output after perturbation as
%$
%\widetilde{\mathcal{Y}} = F(\widetilde{x}_\text{I}, \widetilde{x}_\text{L}; \theta),
%$
%
%where $(\widetilde{x}_\text{I}, \widetilde{x}_\text{L})$ are the images and LiDAR point clouds after perturbation, and $\widetilde{\mathcal{Y}}$ represents the detection outputs.

\noindent
\textbf{Adversarial Objectives:} 
We consider two adversarial objectives: one for false negatives and the other for false positives.
To generate false negative attacks, we follow prior work~\cite{xie2017adversarialdetect,tu2020physically} in attacking object detectors by suppressing all relevant bounding box proposals for the host vehicle. 
A proposal is relevant if its confidence score is greater than 0.1 and it overlaps with the ground-truth bounding box.
The adversarial loss then minimizes the confidence of all candidates:
\begin{align}
\mathcal{L}_\text{adv}^\text{fn} = \sum_{b \in \widetilde{\mathcal{Y}}} -\text{IoU}(b, b^{*}) \log(1 - \texttt{score}(b)),
\end{align}
where IoU denotes the intersection over union operator and $b^{*}$ is the corresponding ground-truth box we aim to attack.

Alternatively, we aim to generate false bounding box proposals that do not overlap with any ground-truth boxes in the scene.
%which we refer to as  false positive generation.
%
The  false positive adversarial loss increases the confidence of the false positive candidates as follows:
\begin{align}
\mathcal{L}_\text{adv}^\text{fp} &= \Sigma_{b \in \widetilde{\mathcal{Y}}_\text{fp}}\log (1 - \texttt{score}(b))\text{ and }\nonumber\\
\widetilde{\mathcal{Y}}_\text{fp} &= \{b|b \in \widetilde{\mathcal{Y}} \text{ and } \forall b^{*} \in \mathcal{Y} \text{ s.t. } \text{IoU}(b, b^{*})=0\},
\end{align}
where $\widetilde{\mathcal{Y}}_\text{fp}$ is a subset of bounding box proposals with no overlap with any ground-truth object bounding boxes.

\begin{figure*}[t]
    \centering
        \includegraphics[width=0.75\linewidth]{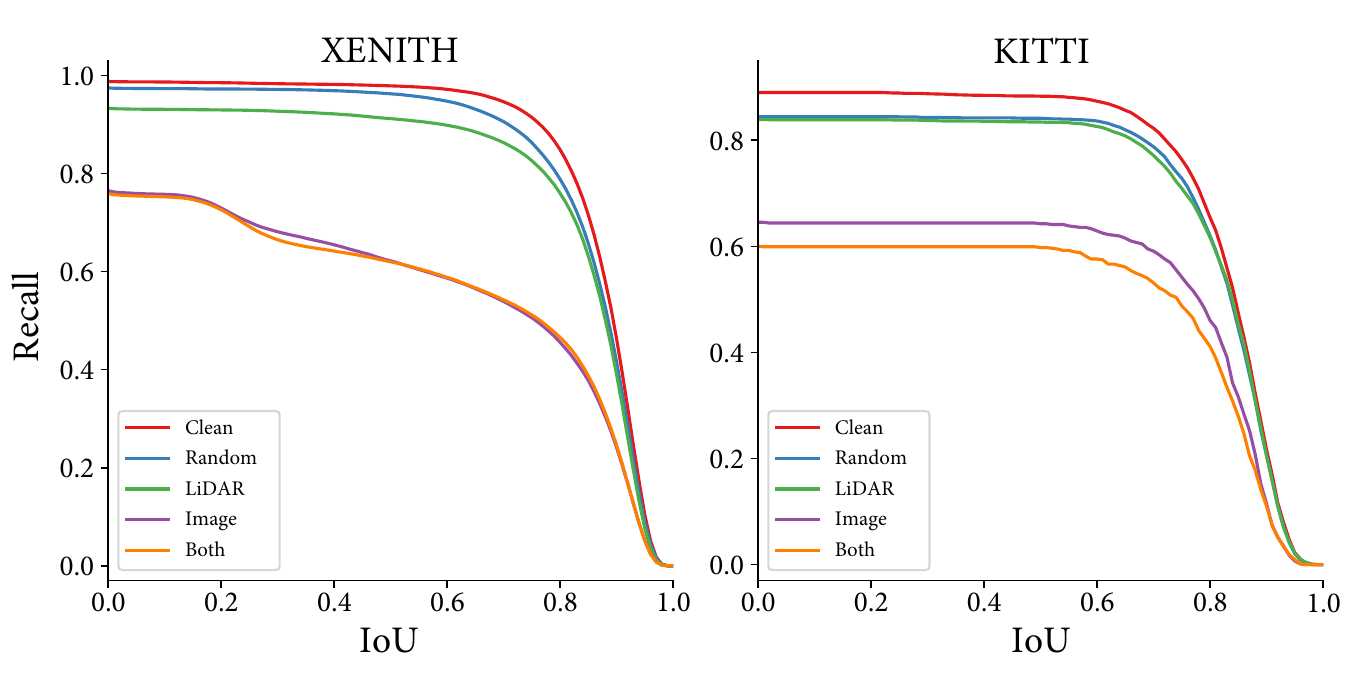}
   \vspace{-2mm}
   \caption{Plot of the host vehicle recall across IoU thresholds. Only attacking LiDAR yields very weak attacks and attacking the image produces significantly stronger perturbations.} 
   \label{fig:iou_rec}
   \vspace{-2mm}
\end{figure*}

\noindent
\textbf{Mesh Regularization:}
Besides the adversarial objective, we use an additional regularization term to encourage realism in the mesh geometry.
Specifically, we use a mesh Laplacian regularizer~\cite{liu2019soft}, which encourages smooth object surface geometries:
$\mathcal{L}_\text{lap} = \sum_{i} {\lVert {\delta}_i \rVert}^2_2$,
with $\delta_i$ as the distance from vertex $v_i \in \mathcal{V}$ to the centroid of its immediate neighbors $\mathcal{N}(i)$: $\delta_i = v_i - \frac{1}{\lVert \mathcal{N}(i) \rVert} \sum_{j \in N(i)} v_j$.
In addition to Laplacian regularization, we also constrain the physical scale of the adversary with an axis-aligned 3D box.
Namely, we require that 
$\norm{\mathcal{V}_j}_{\infty} \leq L_j \text{ for $j \in \{x,y,z\}$}$,
where $L_x$, $L_y$, and $L_z$ represent the box constraints along $xyz$-axis, respectively.

\noindent
\textbf{Learning Input-Agnostic Attacks:}
Overall, our optimization objective can be summarized as 
$\mathcal{L} = \mathcal{L}_\text{adv}^\text{fn} +  \lambda_\text{fp} \mathcal{L}_\text{adv}^\text{fp}  + \lambda_\text{lap} \mathcal{L}_\text{lap},
$
where $\lambda_\text{fp}$ and $\lambda_\text{lap}$ are coefficient that weight the relative importance of the false positive loss term and mesh regularization term.
We employ this objective to optimize the shape and appearance of the inserted object on the entire dataset to generate an input-agnostic adversarial example.
%Therefore, we aim to optimize the expected loss $\mathcal{L}$ across all vehicles in the training distribution. 
%
Therefore, we can denote the optimal adversary as the following equation:
\begin{align}
\mathcal{M}^{*} = \argmin_{\mathcal{M}} \expect_{x, \mathcal{Y}}  \ 
\left[
\mathcal{L}_\text{adv}^\text{fn} +  \lambda_\text{fp} \mathcal{L}_\text{adv}^\text{fp}  + \lambda_\text{lap} \mathcal{L}_\text{lap}
\right].
\end{align}
% 
%Note that we keep the mesh topology or $\mathcal{F}$ unchanged during learning.
%

%We optimize the expected loss by learning a single adversarial mesh with respect to all vehicles in the training set.
%
With our proposed pipeline which is differentiable end-to-end, optimization of the adversarial mesh is done using \textit{projected gradient descent} to respect the $\ell_\infty$ constraints on mesh vertices.
In our experiments, we also conduct attacks targeting a single modality. 
To achieve this, we disable the gradient flow to the untargeted input branch, while we still simulate the mesh into both modalities to maintain physical consistency across image and LiDAR modalities.

\subsection{Multi-sensor Adversarial Robustness}\label{sec:multi-adv-robustness}
Towards defenses against our object insertion attack, we also study defense mechanisms.
Compared to the single-sensor setting, achieving multi-sensor adversarial robustness is even more challenging.
First, each single input modality could be attacked even when the perturbations on the other input sensors are non-adversarial.
Second, adversarial perturbations from each single input modality can interact with each other, which is a unique aspect in the multi-sensor setting.
Thus, we need to deal with not only perturbations at each input modality but also their effect in the fusion layer. % for the multi-sensor adversarial robustness.
\begin{figure*}[t]
	\centering
	\includegraphics[width=0.9\linewidth]{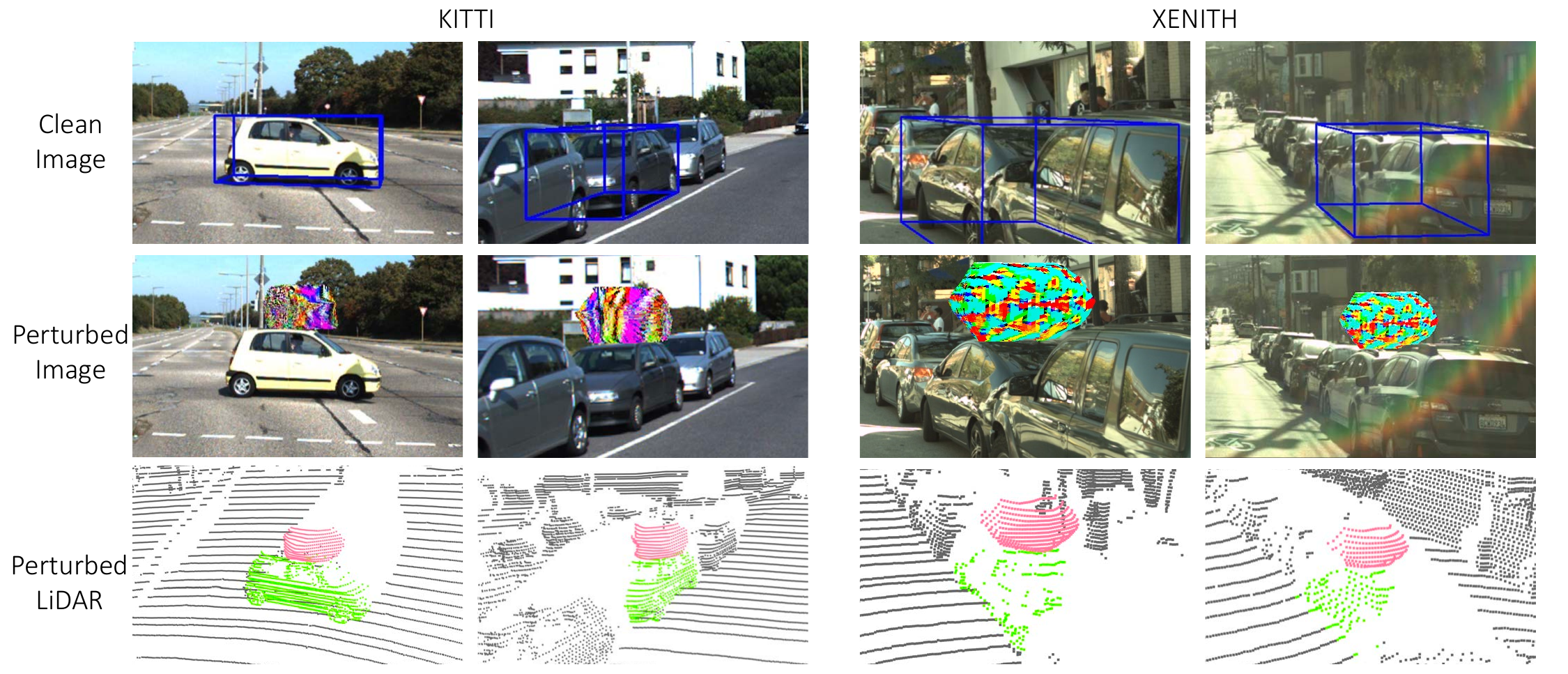}
	\caption{Placing the adversarial mesh on a host vehicle can hide the host vehicle completely from state-of-the-art detectors.The same mesh is used for all vehicles in a dataset as the attack is input-agnostic with respect to the training distribution.}
	\label{fig:qual}
   \vspace{-2mm}
\end{figure*}

We employ adversarial training as it is the most standard and reliable approach to defense. Adversarial training can be formulated as solving for model parameters 
\begin{align}
%\resizebox{0.91\hsize}{!}{%
%$
\theta^{*} = \argmin_{\theta} 
\expect_{x,\mathcal{Y}}  
\left[
\max_{\widetilde{x}_\text{I}, \widetilde{x}_\text{L}}  
%\mathcal{L}_{\text{task}} (F(\widetilde{x}_\text{I}, \widetilde{x}_\text{L}; \theta), \mathcal{Y}) 
\mathcal{L}_{\text{task}} (F(\widetilde{x}; \theta), \mathcal{Y}) 
\right]
%$
%}
\end{align}
which minimize the empirical risk under perturbation. Here $\mathcal{L}_{\text{task}}$ is the loss function used to train the detection model. 
%
%This is achieved by training detection model $F$ against perturbations generated by our threat model $\widetilde{x}$.
This is achieved  by training detection model $F$ against adversarial data $\widetilde{x}$ generated by our attack.
%\xcyan{against adversarial data $\widetilde{x}$ generated by our threat model.}
%\xcyan{Addressing R1 comment``$\tilde{x}$ is adversarial input or threat model? it is unclear what does the author mean by threat model''}
%
While adversarial training is typically performed on image perturbations that are cheap to generate with only few PGD steps~\cite{madry2018towards}, our adversarial example generation is prohibitively expensive for the inner loop of the min-max objective.
Thus, instead of generating a strong adversary from scratch at every iteration, we adopt free adversarial training~\cite{shafahi2019adversarial} and continuously update the same adversary to reduce computation.

%!TEX root = main.tex
\section{Experimental Evaluations}
\label{sec:exp}

In this section, we first describe %in Sec.~\ref{sec:exp-setting} 
the our datasets, attack protocols, and evaluation metrics. More details on experiment setting are provided in the supplementary material.
We then conduct experiments on the multi-sensor detection model MMF~\cite{liang2019multi} and present our empirical findings for \textit{white-box} attacks on each dataset and the \textit{black-box} transfer attacks across datasets.
Finally, %in Sec.~\ref{sec:exp-robustness}, 
we explore several defense mechanisms to achieve a more robust multi-sensor object detector.

\subsection{Experimental Setting}\label{sec:exp-setting}
\paragraph{Datasets:} 
We conduct our experiments on two self-driving datasets: KITTI~\cite{Geiger2012CVPR} and \ourdataset{}.  \ourdataset{} is collected  by having a fleet of self-driving cars drive around cities in North America. 
%
%For both datasets,
Snippets are captured in the daytime and detection is performed on objects within 70 meters forward and 40 meters to the left and right of the ego-car.
%using the front camera image and the LiDAR point clouds.
%
%%KITTI images are provided at a $370 \times 1224$ resolution and \ourdataset{} images are provided at a $320 \times 2048$ resolution. 
%
%%We treat each vehicle in a frame as a separate sample and specifically require the adversarial mesh inserted on the rooftop of a host vehicle to be at least 70\% visible from the image.
%
Each vehicle in a frame is considered a separate sample and we have 12,284 vehicles in KITTI and 77,818 in \ourdataset{}.
%% which we further split into a training and test set at a 7:3 ratio. 

%\paragraph{Multi-sensor Object Detection:} 
%We adopt the multi-sensor network architecture of MMF~\cite{liang2019multi} to perform object detection in bird's eye view (BEV) on KITTI dataset.
%
%On \ourdataset{}, we use a slightly different architecture which has been tuned for  improved performance (see our supplementary for details).
%
%The multi-sensor detector has two separate branches to extract features from images and LiDAR, in which LiDAR points are voxelized and processed as a BEV image with 2D convolutions while images are processed with a ResNet backbone~\cite{he2016deep}.
%
%For sensor fusion, each voxel in BEV performs K-nearest neighbour (KNN) to sample close-by LiDAR points.
%
%We follow the implementation of prior work~\cite{liang2019multi} and set K=1 in our fusion modules.

\paragraph{Metrics:} 
Following prior work on object insertion attacks~\cite{tu2020physically}, we evaluate how often the host vehicle ``disappears'' by measuring its \textit{recall} across various IoU thresholds. 
%we aim to generate false negatives and make the host vehicle ``disappear'' from the detector. 
%\raquel{mmm this is not what you wrote in the attack model section}\raquel{just remove the "mainly"}
%
For a scalar metric, we evaluate the \textit{false negative attack success rate (FN ASR)} as the percentage of host vehicles detected before perturbation that are undetected after perturbation.
We consider a vehicle detected if there exists an output bounding box having greater than 0.7 IoU with the vehicle.
%
%In addition, to generate missed detections on the host vehicle, false positives may also be generated. 
%
On the other hand, we consider an output bounding box a false positive if its maximum IoU with any  ground truth box is less 0.3 and it does not overlap with any detection produced before perturbation. 
%
%Furthermore, we do not count boxes which overlap with the clean detections or the host vehicle.
%
We evaluate the \textit{false positive attack success rate (FP ASR)} as the percentage of attacks which generate at least one false positive. 
Finally, the overall \textit{attack success rate (ASR)} is the percentage of attacks which successfully create a false positive or false negative.

\paragraph{Implementation Details:}
The adversarial mesh is initialized as an icosphere with $N=162$ vertices and $M=320$ faces.
Per-face textures are parameterized using a texture atlas with a $5 \times 5$ texture resolution for each face.
During optimization, we set $\lambda_{\text{lap}} = 0.001$, $\lambda_{\text{fp}}=1$, 
and further constrain the scale of the mesh with an axis-aligned 3D box where the $x$ and $y$ coordinates are bounded by $L_x=L_y=0.8m$ and the $z$ coordinate is bounded by $L_z=0.5m$. 
We use Adam~\cite{kingma2014adam} to optimize the mesh parameters with a learning rate of $0.004$ for textures and $0.001$ for vertex coordinates. 
To target either LiDAR or image branch in isolation, we disable gradient flow to the other branch during the backward pass to the adversary.

\subsection{Universal Adversarial Attacks}\label{sec:exp-attack}

\paragraph{Hiding Host Vehicle:}
We evaluate the drop in recall in detecting the host vehicle, as missed detections can lead to colliding with unseen objects which is the most dangerous outcome.
We sweep IoU thresholds and visualize the IoU-recall curve in \figref{fig:iou_rec}. 
First, inserting a mesh with randomized shape and appearance has little impact on the detector.
On the other hand, an adversarial mesh generated by perturbing both input modalities leads to a significant drop in recall. 
Moreover, we perturb the LiDAR and image inputs in isolation and find that targeting the LiDAR inputs alone yields very weak attacks. 
Meanwhile, targeting the image alone is almost as strong as perturbing both modalities. 
Therefore, image inputs are significantly less robust to the proposed attack.

\paragraph{Attack Success Rates:} 
\begin{wrapfigure}{R}{0.45\textwidth}
\vspace{-\baselineskip}
\caption{Size of adversary box constraint vs attack success rate.}
%Size of the adversary is varied as sweep the $\ell_{\infty}$ constraint $L$ for the vertex coordinates. Larger meshes lead to stronger attacks.}
\vspace{-1.5mm}
\includegraphics[width=1.0\linewidth, trim={0 3.5mm 0 2mm}, clip]{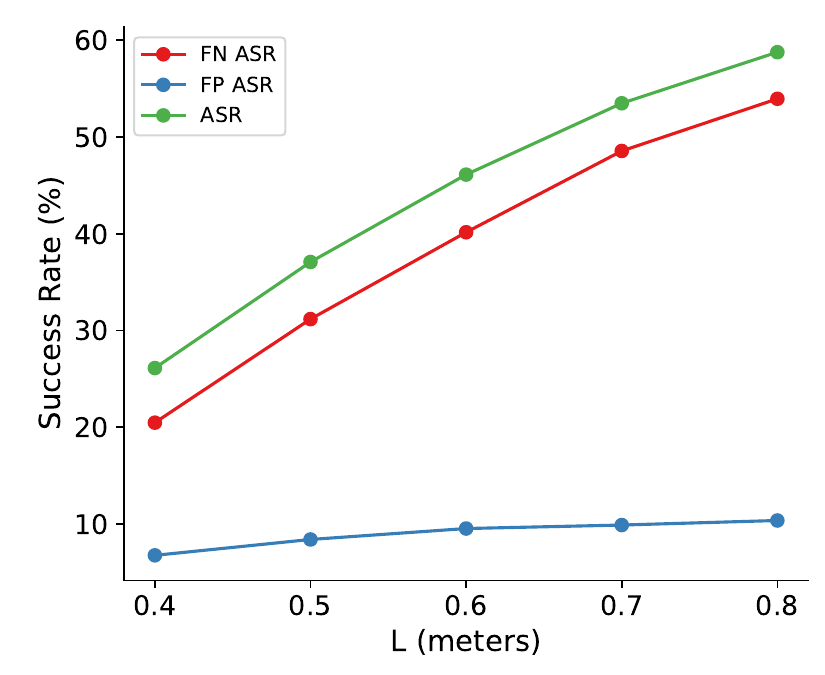}
\vspace{-8.5mm}
\label{fig:sz_sweep}
\end{wrapfigure}

In \tabref{tab:kitti_asr} and \tabref{tab:xenon_asr}, we further analyze results in terms of the attack success rates. 
Again, we consider meshes with randomly generated geometry and texture as a baseline.
We observe similar trends of image features being significantly more vulnerable. 
In addition to missed detections, the adversarial mesh is able to attack the detector through generating false positive proposals of objects that do not exist.
Furthermore, we compare against prior work~\cite{tu2020physically} which attacks a LiDAR-only detector.
In this case, incorporating image inputs boosts robustness to LiDAR attacks at the cost of being more vulnerable to multimodal attacks.
\paragraph{Adversary Size}
Furthermore, to understand how much the size of the adversary affects the strength of the attack, we vary the size of the box constraints on the vertex parameters. Here we sweep $L_x = L_y = L_z = L$ and and measure the attack success rates. Results are shown in \figref{fig:sz_sweep}. As expected, the attack becomes stronger as the $\ell_{\infty}$ constraints on vertex coordinates are relaxed. 
%The improvements are more noticeable in the false negative success rates.

%\setlength{\tabcolsep}{4pt}
\begin{table}
    \centering
    \setlength{\tabcolsep}{3pt}
    \begin{minipage}[b]{0.48\linewidth}
    \centering
    \resizebox{1.\linewidth}{!}{
    \begin{tabular}{@{} l|l|cccr  @{}}
        \toprule 
         Detector & Attack & FN ASR & FP ASR & ASR \\
         \midrule         
          LiDAR & LiDAR~\cite{tu2020physically}  & 31.85\%  & 4.84\%  & 33.23\%\\
          \midrule      
         \multirow{4}{*}{\shortstack[c]{LiDAR \\ +\\Image}} & Random &  5.68\% & 2.01\% & 7.64\% \\ 
         & LiDAR      & 7.99\% & 2.36\%  & 10.11\%  \\
         & Image      & 26.06\% & 3.40\%  & 28.43\%  \\
         & Both   & \textbf{32.76\%}  & \textbf{4.38\%}  & \textbf{34.68\%}   \\
        \bottomrule
    \end{tabular}
    }
    \caption{Attacks on KITTI. Also comparing with random meshes and a LiDAR only model.}
    \label{tab:kitti_asr}
    \end{minipage}\hfill
    \begin{minipage}[b]{0.48\linewidth}
    \centering
	\resizebox{1.\linewidth}{!}{
		\begin{tabular}{@{} l|l|cccr  @{}}
			\toprule 
			Detector & Attack & FN ASR & FP ASR & ASR \\
			\midrule                         
			LiDAR & LiDAR~\cite{tu2020physically}  & 23.80\% & 10.70\% & 32.60\% \\
			\midrule
			\multirow{4}{*}{\shortstack[c]{LiDAR \\ +\\Image}} & Random & 5.06\% & 4.15\% & 9.17\% \\ 
			& LiDAR & 9.52\% & 6.21\% & 15.33\% \\
			& Image & 42.81\% & 10.78\% & 49.59\% \\
			& Both & \textbf{43.15\%} & \textbf{ 11.77\%} & \textbf{49.76\%}   \\
			\bottomrule
		\end{tabular}
	}
	\caption{Attacks on \ourdataset{}. Also comparing with random meshes and a LiDAR only model.}
	\label{tab:xenon_asr}
    \end{minipage}
\end{table}

\begin{figure*}
	\centering
	\includegraphics[width=0.99\linewidth]{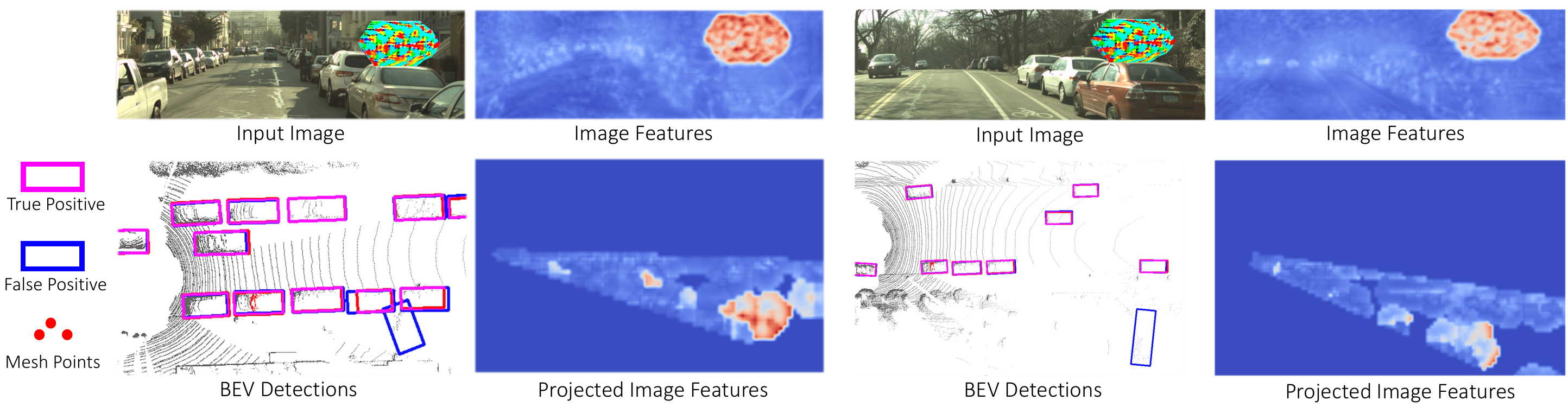}
    \vspace{-2mm}
	\caption{
    Visualization of the attack producing distant false positives due to the camera's perspective projection, as well as the corrupted image features in the image plane and projected in 3D.
    }
    \vspace{-3mm}
	\label{fig:fp_qual}
\end{figure*}

\paragraph{Qualitative Examples:}
Qualitative examples are shown in \figref{fig:qual}. 
First, the detector fails to detect host vehicles with the adversarial mesh on its rooftop.
We show detections in the image rather than LiDAR for ease of viewing. 
Note that the same adversarial mesh is used for all demonstrations, as the attack is agnostic to the host vehicle and environment.
Furthermore, we show in \figref{fig:fp_qual} that our adversarial mesh generates false positives at very distant locations. 
Here, detections are visualized in BEV since distant objects appear too small in the image.
Additionally, we visualize image features in the image plane and the visual cone of projected image features into 3D, showing that long-range false positives are caused by strong image features dominating after fusion.

\paragraph{Black-box Transfer Attacks:}
We conduct transfer attacks across datasets and show results in \tabref{tab:attack_transfer}. 
Overall, our transfer attack on the target dataset is stronger than attacking only the LiDAR input modality on the source dataset, especially from \ourdataset{} to KITTI.
On the other hand, the transferability is probably lowered by the image resolution and hardware, which is beyond the scope of our paper but an interesting future direction to explore.

\begin{table}
\centering
\begin{minipage}[t]{0.44\linewidth}
    \setlength{\tabcolsep}{2.5pt}
    \centering
    \resizebox{1.\linewidth}{!}{
    \begin{tabular}{@{} l|l|cccr  @{}}
        \toprule 
         Source & Target & FN ASR & FP ASR & ASR \\
     \midrule                         
         \multirow{2}{*}{KITTI} & KITTI & 32.76\%  & 4.38\%  &  34.68\%   \\
         & \ourdataset{}  & 14.20\%  & 2.86\%  & 16.88\%  \\
         \midrule
         \multirow{2}{*}{\ourdataset{}}&  KITTI  &  12.64\%  & 6.12\%  & 18.22\%  \\
         & \ourdataset{}  & 43.15\%    & 11.77\%  & 49.76\%    \\
        \bottomrule
    \end{tabular}
    }
    \caption{Transfer - KITTI \& \ourdataset{}}
    %\caption{Black box transfer attack results between KITTI and \ourdataset{}. We observe some transferability even between two datasets collected with different sensor hardware and in different geographic locations.}
    \label{tab:attack_transfer}

\end{minipage}\hfill
\begin{minipage}[t]{0.51\linewidth}
\setlength{\tabcolsep}{2pt}
    \centering
    \resizebox{1.\linewidth}{!}{
    \begin{tabular}{@{} l|ccc | cr  @{}}
        \toprule 
         Defense & FN ASR & FP ASR & ASR & AP(clean) \\
         \midrule                         
         None & 43.15\%  & 11.77\%  & 49.76\%  & \textbf{84.64\%}  \\
         \midrule
         JPEG~\cite{dziugaite2016study} & 43.19\% & 9.45\% & 49.60\% &  84.52\%   \\
         Adv Train~\cite{shafahi2019adversarial} & 7.83\% & 8.29\% & 14.97\%  & 84.16\%   \\
         Adv FD~\cite{xie2019feature}  & \textbf{3.57\%} & \textbf{7.53\%} & \textbf{10.82\%}  & 83.97\%  \\
        \bottomrule
    \end{tabular}
    }
    \caption{Defense results on \ourdataset{}}
    %\caption{Defense results on \ourdataset{}. Adversarial training significantly boosts robustness and is improved with feature denoising. \texttt{JPEG} compression is not effective. However, these methods still struggle in defending false positives.}
    \label{tab:defense}
\end{minipage}
\vspace{-8mm}
\end{table}

\vspace{-1mm}
\subsection{Improving Robustness}\label{sec:exp-robustness}
\vspace{-1mm}

\paragraph{Attacks Against Defense Methods:}
As empirical findings suggest that the image feature is more vulnerable, we first employ an existing image-based defense method that removes high-frequency component through~\texttt{JPEG} compression~\cite{dziugaite2016study}.
In addition, we conduct adversarial training against the attacker.
Since generating a strong adversary is extremely expensive due to the simulation pipeline, we employ a strategy similar to Free Adversarial Training~\cite{shafahi2019adversarial} and reuse past perturbations by continuously updating the same adversarial object. 
Specifically, we perform 5 updates to the adversary per one update to the model.
We combine the feature denoising~\cite{xie2019feature} with the adversarial training to further enhance robustness against image perturbations in particular.
We report the success rates as well as the average precision (AP) at 0.7 IoU to study the trade-off between adversarial robustness and performance on benign data~\cite{tsipras2018robustness}.

As shown in \tabref{tab:defense}, we find \texttt{JPEG} compression is very ineffective as defense.
We hypothesize this is because the  input-agnostic adversary is rendered at various different poses during training and therefore do not rely on high-frequency signals that are removed by \texttt{JPEG} compression.
In comparison, our adversarial training effectively reduces the overall attack success rate from 49.76\% to 14.97\%, while dropping AP by 0.5\%. 
Finally, adding non-local mean blocks after every residual block in the image processing backbone further improves robustness by reducing the ASR by 5\%.

\paragraph{Discussions and Future Work:} 
While adversarial training methods are effective, they are specific to a specific threat model and may not generalize to unseen perturbations.
A more threat-agnostic mechanism like more robust sensor fusion would bring more robustness in general.  
Furthermore, adversarial defense methods are only effective at recovering the missed detections, but struggle to detect false positives.
We believe this is because distant false positives shown in \figref{fig:fp_qual} are only partially due to vulnerabilities to adversarial perturbations.
In fact, such examples exploit erroneous associations between objects that are distant in 3D.
Specifically, the mapping between a mesh pixel and a LiDAR point far away from the mesh enables such attacks.    
These false associations can easily occur if the assigned pixel for each LiDAR points is shifted by a few pixels, since objects which are far apart in 3D may appear very close in 2D. 
We identify two reasons how this can occur in practice.
First, due to the receptive field of DNN activations, an adversarial object can influence pixels outside its physical boundaries.
Second, while LiDAR sweeps are collected in a continuous fashion with a rotating sensor, images are captured instantaneously at regular intervals. 
Consequently, the camera extrinsics used for projection become outdated for LIDAR points captured before and after the image. 
Thus, to achieve more robust sensor fusion for images and LiDAR, fusion modules must reason about 3D geometry, contextual information, and temporal information of LiDAR points to generate mappings between image pixels and LiDAR points more intelligently.
We hope these findings will inspire future work towards more robust sensor fusion methods.

%!TEX root = main.tex
\section{Conclusions}

Our work investigates practical adversarial attack against mulit-sensor detection models in self-driving to understand how consuming multiple input modalities affect adversarial robustness.
Compared to existing attacks against multimodal detectors, our object insertion attack is more threatening in practice as we generate input-agnostic and physically realizable adversarial perturbations.
%
% The proposed method is capable of hiding existing vehicles and generating false detections at the same time.
%
Our experiments % evaluations on KITTI and \ourdataset{} 
reveal that vulnerabilities of multi-sensor object detectors are primarily due to non-robust image features.
While adversarial training can effectively recover the missed detections, we find it still struggles to detect 
false positives without a deeper reasoning about 3D geometry in feature fusion.
We believe this work would open up new research opportunities and challenges in the field of multi-sensor robustness.

% The maximum paper length is 8 pages excluding references and acknowledgements, and 10 pages including references and acknowledgements

\clearpage
% The acknowledgments are automatically included only in the final and preprint versions of the paper.
%\acknowledgments{If a paper is accepted, the final camera-ready version will (and probably should) include acknowledgments. All acknowledgments go at the end of the paper, including thanks to reviewers who gave useful comments, to colleagues who contributed to the ideas, and to funding agencies and corporate sponsors that provided financial support.}

%===============================================================================

% no \bibliographystyle is required, since the corl style is automatically used.
\bibliography{bib}  % .bib

\end{document}